\documentclass[10pt, conference, letterpaper]{IEEEtran}
\IEEEoverridecommandlockouts

\usepackage{cite}
\usepackage{amsmath,amssymb,amsfonts}
\usepackage{amsthm}
\usepackage{mathtools}
\usepackage{enumitem}
\usepackage{algorithmic}
\usepackage{graphicx}
\usepackage{wrapfig}
\usepackage{diagbox}
\usepackage{textcomp}
\usepackage[table]{xcolor}
\usepackage{tcolorbox}
\def\BibTeX{{\rm B\kern-.05em{\sc i\kern-.025em b}\kern-.08em
    T\kern-.1667em\lower.7ex\hbox{E}\kern-.125emX}}

\usepackage{tabularx}
\usepackage{booktabs}
\usepackage{threeparttable}
\usepackage{makecell}
\usepackage{pifont}
\usepackage{hyperref}
\usepackage{tikz}
\usepackage{inconsolata}
\usepackage[linesnumbered,ruled,vlined]{algorithm2e}
\SetCommentSty{\fontsize{5}{5}\selectfont} 
\SetKwComment{tcc}{$\triangleright\,$}{}

\SetCommentSty{mycommfont}
\SetKwInput{KwData}{Require}
\SetKwInput{KwResult}{Result}

\definecolor{mygreen}{RGB}{11,141,10}
\definecolor{myred}{RGB}{223,68,52}
\definecolor{myblue}{RGB}{70,130,180}
\definecolor{mydeepblue}{RGB}{65,105,225}
\definecolor{myviolet}{RGB}{97,0,138}
\definecolor{myburgundy}{RGB}{110,10,30}
\definecolor{myblue2}{RGB}{0,105,148}
\definecolor{grayhighlight}{RGB}{250,250,227}
\definecolor{deepblue}{rgb}{0.0, 0.0, 0.55} %

\hypersetup{
    colorlinks=true,      %
    linkcolor=deepblue,       %
    citecolor=myviolet,      %
    filecolor=magenta,    %
    urlcolor=black,        %
    pdftitle={Document Title},    %
    pdfauthor={Author Name},      %
    pdfsubject={Subject of the document}, %
    pdfkeywords={keyword1, keyword2, keyword3}, %
    bookmarksnumbered=true, %
    bookmarksopen=true,     %
    pdfpagemode=UseOutlines,%
    pdfstartview=Fit       %
}

\usepackage{stmaryrd}

\newcommand{\ie}{{\em i.e.}}

\newcommand{\aka}{{\em a.k.a.}}

\newcommand{\etal}{{\em etal.}}

\usepackage{siunitx} %

\usepackage{cryptocode}

\usepackage{CJKutf8}
\begin{document}
\begin{CJK*}{UTF8}{gbsn}

\title{Exploring Chinese Humor Generation: A Study on Two-part Allegorical Sayings\\
}

\author{
  \IEEEauthorblockN{Rongwu Xu}
  \IEEEauthorblockA{IIIS, Tsinghua University}
  \IEEEauthorblockA{\texttt{xrw22@mails.tsinghua.edu.cn}}
}

\maketitle

\begin{abstract}
Humor, a culturally nuanced aspect of human language, poses challenges for computational understanding and generation, especially in Chinese humor, which remains relatively unexplored in the NLP community.
This paper investigates the capability of state-of-the-art language models to comprehend and generate Chinese humor, specifically focusing on training them to create allegorical sayings. We employ two prominent training methods: fine-tuning a medium-sized language model and prompting a large one.
Our novel fine-tuning approach incorporates fused Pinyin embeddings to consider homophones and employs contrastive learning with synthetic hard negatives to distinguish humor elements. Human-annotated results show that these models can generate humorous allegorical sayings, with prompting proving to be a practical and effective method. However, there is still room for improvement in generating allegorical sayings that match human creativity.

\end{abstract}

\section{Introduction}
\label{sec:intro}

Recent advancements in natural language processing (NLP) have been largely driven by pre-trained transformer-based LMs like BERT~\cite{devlin2018bert} and RoBERTa~\cite{liu2019roberta}, excelling in natural language understanding (NLU) tasks such as sentiment analysis and named entity recognition (NER). Seq2Seq LMs like GPT~\cite{radford2018improving} and T5~\cite{raffel2020exploring} have shown impressive natural language generation (NLG) capabilities. The evolution of large language models (LLMs) like those in~\cite{brown2020language, openai2023gpt4,touvron2023llama} is a significant step towards artificial general intelligence~\cite{bubeck2023sparks}, demonstrating strong few-shot and zero-shot in-context learning (ICL) abilities~\cite{brown2020language,dong2022survey}.

\begin{figure}
    \centering
    \includegraphics[width=0.8\linewidth]{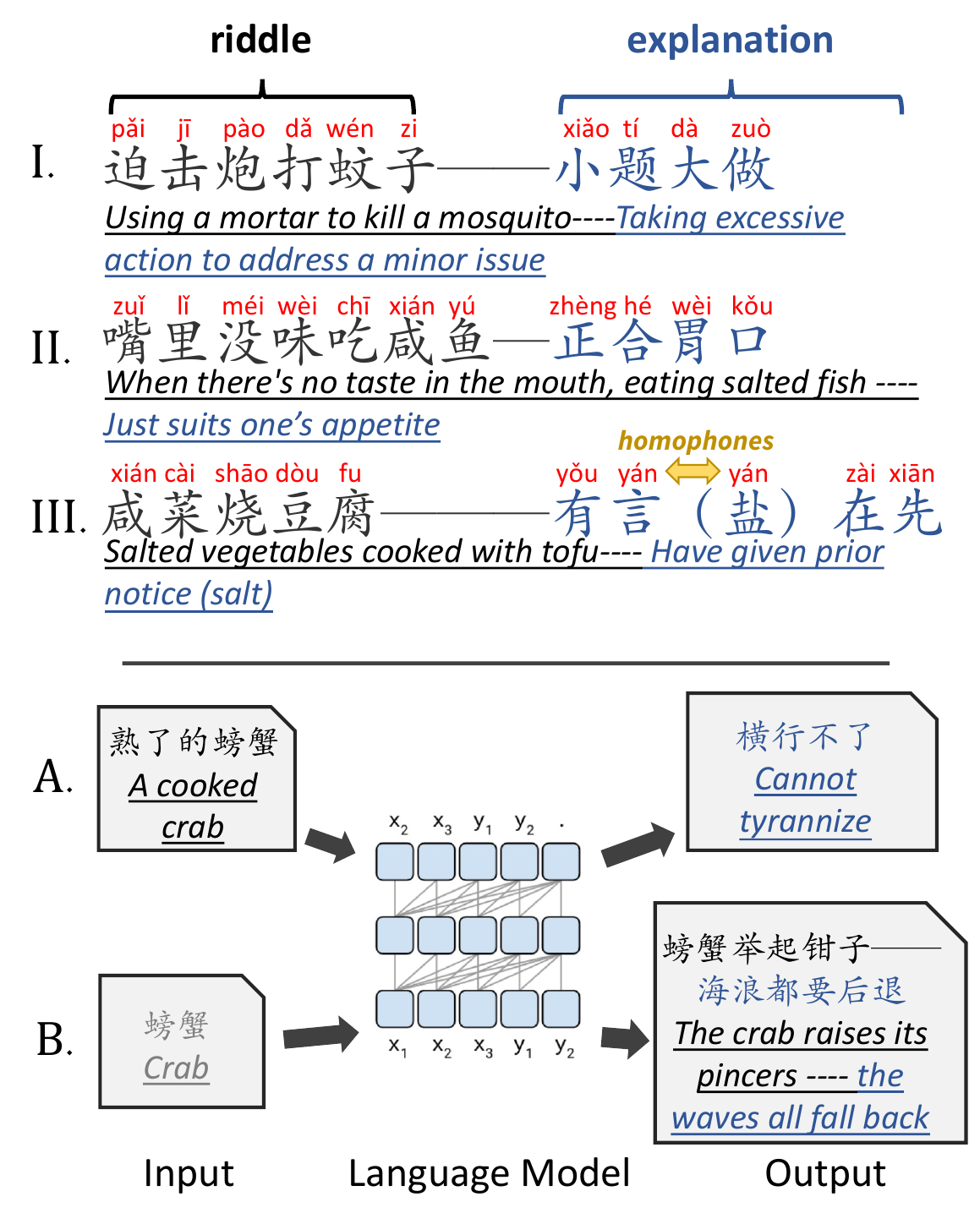}
    \caption{(\textbf{Top}) Examples of Chinese two-part allegorical sayings. The ``riddle'' parts are marked in black with the ``explanation'' in blue. Pinyin and translation are noted in red and underlined texts, respectively.
    The three representative types of allegorical sayings are: \textbf{I.} \emph{Metaphor/Allegory}-based saying,
    \textbf{II.} \emph{Reasoning}-based saying, and
    \textbf{III.} \emph{Homophone}-based saying. (\textbf{Bottom}) Two generation tasks: \textbf{A}. Completion of explanations on given riddles and \textbf{B}. Generation of complete sayings on given entities that act as topics.}
    \label{fig:demo}
\end{figure}

Humor, a complex aspect of human communication, poses challenges in computational understanding and generation within NLP, despite its deep theoretical roots~\cite{bardon2005philosophy, humorisgoodlife}. Transformer-based LMs have propelled research in humor recognition and generation, particularly in English~\cite{mao2019bert, weller2019humor, annamoradnejad2020colbert, weller2020can, zhang2020let}. However, computational humor studies in non-English languages remain limited due to data imbalance~\cite{winters2020dutch,chen2023can}.
While humor is a universal aspect of human experience, its expression varies significantly among different languages and cultures.
Chinese humor focuses on wordplay and homophones, less on direct jokes. It often uses satire and paradoxical statements to reflect traditional wisdom and societal commentary. Allegorical stories and concise anecdotes are also common, featuring witty and ironic scenarios with historical characters.
In this paper, we concentrate on two-part allegorical sayings (Chinese: 歇后语\footnote{It can be literally translated to ``saying with a rest before the second part''. ``语'' means ``saying'', and ``歇后'' means ``take a rest before doing something''.}, Pinyin\footnote{Pinyin is the Romanization of the Chinese characters based on their pronunciation.}: xiē hòu yǔ), which demonstrate a comprehensive blend of these key elements~\cite{lai2008understanding}. 
A two-part allegorical saying starts with a familiar scenario (the first part, referred to as the ``riddle'') and concludes with an indicative, unexpected, and often humorous twist (the second part, referred to as the ``explanation''), as illustrated in~\autoref{fig:demo} (Top).
In addition to their humorous effect, allegorical sayings have the power to make those who employ them appear more knowledgeable and articulate.
In the remaining sessions of this paper, we will abbreviate the term ``Chinese two-part allegorical saying'' to ``allegorical saying''.

In this paper, \textbf{we delve into SOTA LMs' ability to comprehend and generate Chinese humor through the lens of \emph{training them to generate allegorical sayings}.}
To achieve our objective, we adopt two prominent training paradigms: (i) fine-tune a medium-sized LM after pre-train, and (ii) prompt a powerful LLM.
In approach (i), we propose a three-stage training procedure.
Initially, we continue to pre-train the LM on large-scale Chinese corpora sourced from the web. The LM is augmented with learnable Pinyin embeddings to assist the model in considering homophones.
Subsequently, we introduce a specialized contrastive learning method to further train the LM to distinguish correctly paired allegorical sayings from synthetic hard negative samples.   
In the final stage, we fine-tune the LM on a smaller allegorical saying dataset, with labeled homophones in the training samples.
In approach (ii), we utilize zero-shot and few-shot prompting techniques to assess the saying generation capabilities of off-the-shelf LLMs.

In our research, we investigate two generation tasks (see~\autoref{fig:demo} (Bottom)): (A) Completing riddle explanations, and (B) Generating complete sayings based on given entities. 
Due to the absence of suitable evaluation metrics, in this context, we engaged proficient Chinese annotators to score the generated sayings based on two criteria: coherency and humor, on a scale of 1 to 3.
Based on our scoring indicators, in the completion task, our fine-tuned mT5-small model achieved coherency/humor scores of $2.13/1.59$, while the ChatGPT model with few-shot prompting obtained scores of $2.45/1.32$. In comparison, the gold human-written samples received the highest scores of $2.89/2.06$.
Our findings suggest:
(i) LMs can generate allegorical sayings infused with Chinese humor;
(ii) Prompting an LLM proves to be a convenient and effective approach, yielding results comparable to the fine-tuned LM;
(iii) A noticeable gap still exists in the quality and diversity of computationally generated sayings compared to those created through human ingenuity.

\noindent \textbf{Contribution.} Our contributions are fourfold.
\begin{itemize}
    \item We pioneer the investigation into LMs' capabilities in Chinese humor generation, filling a gap in the existing research landscape. To the best of our knowledge, this is the first exploration in generating allegorical sayings.
    \item We design a seq2seq Pinyin-augmented LM that incorporates additional Pinyin embeddings as input. This LM is versatile and can be applied to various Chinese NLG tasks where Pinyin plays an essential role.
    \item We propose a three-stage fine-tuning methodology, notable for its inclusion of contrastive learning using synthetic hard negatives. This approach enhances the LM's proficiency in generating allegorical sayings.
    \item We employ human annotators to carefully examine and score the generated samples, adding a valuable layer of analysis to our experiments.
\end{itemize}

\noindent\textbf{Roadmap.} The rest of this paper is organized as follows. 
We briefly introduce background in ~\autoref{sec:background}.
\autoref{sec:finetuning} and~\autoref{sec:prompting} introduce our proposed three-stage fine-tuning approach and the prompting approach for allegorical saying generation, respectively. 
We present evaluation results and findings in~\autoref{sec:experiments}.
Related work are reviewed in~\autoref{sec:related}.
\autoref{sec:conclusion} concludes our paper.
\section{Background}
\label{sec:background}

Chinese allegorical sayings employ multiple linguistic techniques to add humor, predominantly in three ways, as shown in~\autoref{fig:demo}. 
Firstly, \emph{metaphors}, usually found in the explanation part, often provide a critical or satirical interpretation of the scene described in the riddle; however, some sayings have metaphors embedded directly in the riddle itself.
Secondly, \emph{logical reasoning} interprets the scenario given in the riddle, typically highlighting the absurdity of futile or ineffective actions.
Thirdly, the use of \emph{homophones}, almost always appearing in the explanation part, cleverly shifts the initial interpretation of the riddle, leading to an unexpected twist.
Creating an allegorical saying demands a high degree of creativity and skill. It involves the artful application of literary techniques to construct a phrase that is not only humorous but also imbued with philosophical insights. These sayings are akin to compact narratives, presenting themselves as riddles steeped in wisdom and jokes intricately laced with wordplay. This task challenges the creator to strike a balance between brevity and depth.

\section{A Pre-training and Fine-tuning Approach}
\label{sec:finetuning}

We begin by introducing the architecture of our Pinyin-augmented model, followed by a detailed description of our proposed three-stage training procedure.

\begin{figure*}
    \centering
    \includegraphics[width=0.95\linewidth]{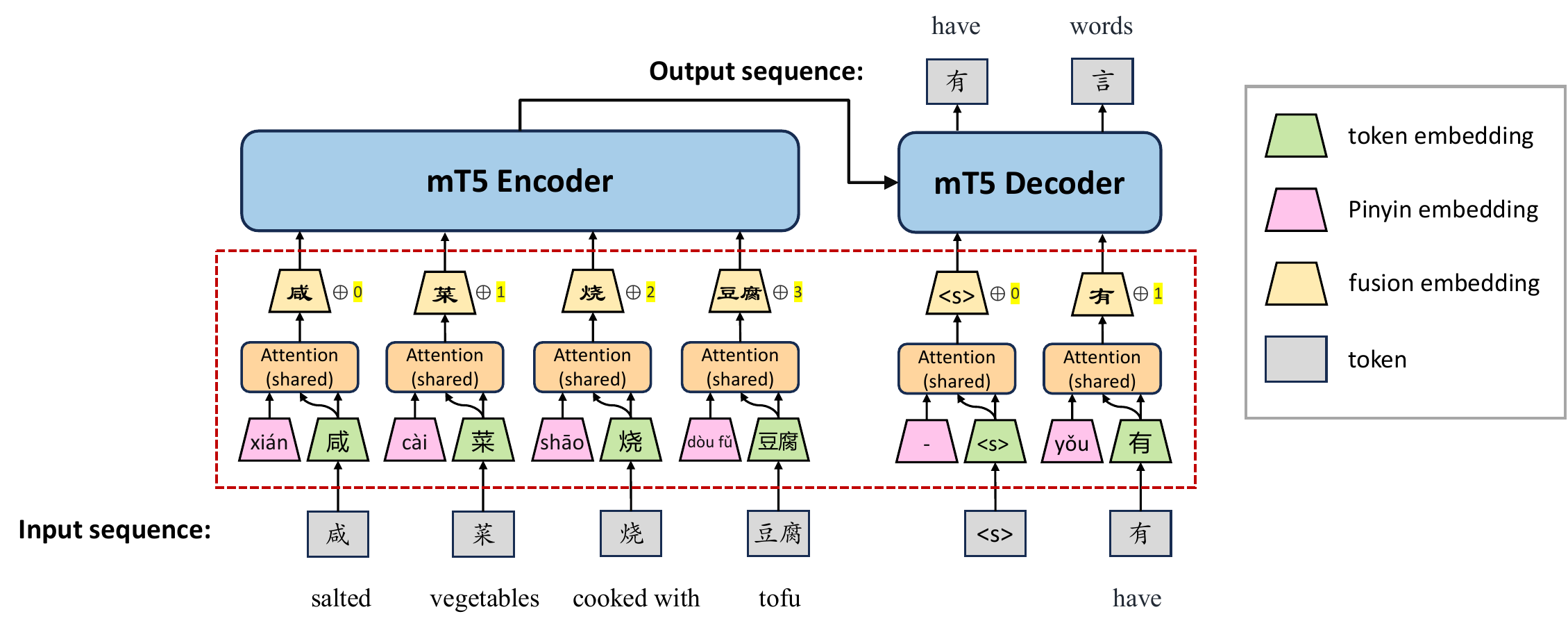}
    \caption{An overview of our PmT5 language model, which includes Pinyin inputs. The modifications are highlighted within the red dashed frame. Our enhancements consist of two main components: (a) a Pinyin embedding layer that converts Pinyin into vectors of size $d_{model}$, and (b) a multi-head attention layer to fuse the Pinyin embedding with token embeddings, resulting in a combined fusion embedding, also of size $d_{model}$. (In the original T5/mT5 model, where token embeddings are directly added with positional encodings to serve as the encoder's input.)}
    \label{fig:model}
\end{figure*}

\subsection{Model Architecture}

In our sequence-to-sequence (seq2seq) generation task, we chose the mT5 LM (T5's multilingual version) as our backbone for the fine-tuning approach. As highlighted in~\cite{xue2020mt5}, due to its outstanding performance in various seq2seq tasks and its ability to understand and generate content in multiple languages, including Chinese.

\noindent \textbf{Incorporating Pinyin inputs.}
The original mT5 model is not specifically tailored for Chinese natural language generation (NLG) tasks that heavily emphasize the Pinyins of Chinese characters. This aspect is crucial in handling allegorical sayings. In addition, as illustrated in~\cite{sun2021chinesebert}, Pinyin provides an additional semantic layer beyond its token embedding that the same Chinese character(s) can yield different meanings based on their pronunciation.
For instance, the character ``乐'' conveys the semantic of ``happy'' when pronounced as ``lè'' (as in ``快乐'', meaning ``joy'') and ``music'' when pronounced as ``yuè'' (as in ``乐器'', meaning ``musical instrument'').

Following ChineseBERT~\cite{sun2021chinesebert}, we also add another embedding input for processing the corresponding Pinyin of a token. 
Our PmT5 (\underline{P}inyin-augmented \underline{mT5}) model's structure is illustrated in~\autoref{fig:model}. 
We make the following modifications to the original mT5 model:
\begin{enumerate}
    \item \emph{Pinyin embedding}: The Pinyin embedding converts the corresponding Pinyins of one or more characters in a token into an embedding of dimension $d_{model}$, where $d_{model}$ represents the dimensions of sub-layer outputs~\cite{vaswani2017attention}.
    Similar to ChineseBert~\cite{sun2021chinesebert}, we apply a TextCNN~\cite{kim-2014-convolutional} with a kernel size of 2 for processing the Romanized Pinyin sequences using the Python \texttt{pypinyin} package\footnote{\url{https://pypi.org/project/pypinyin/}}. 
    We choose to exclude the four diacritics for tonal distinctions in the input of Pinyin embedding.
    \item \emph{Multi-head attention to fuse Pinyin and token}: We employ a standard multi-head attention module~\cite{vaswani2017attention} denoted as $\text{MultiHead}(Q,K,V)$ to effectively fuse the Pinyin and token embeddings with Pinyin embedding $e_{\text{pinyin}}$ serves as the query $Q$, while the token embedding $e_{\text{token}}$ functions as both the key $K$ and value $V$:
    \begin{equation}
        e_{\text{fusion}} = \text{MultiHead}(e_{\text{pinyin}}, e_{\text{token}}, e_{\text{token}}).
    \end{equation}
    The resultant fusion embedding $e_{\text{fusion}}$ is then added with positional encoding before being fed to the encoder.
\end{enumerate}
Please note that we did not simply borrow the add-ons of ChineseBERT~\cite{sun2021chinesebert} from BERT to mT5, the key distinction is that we incorporate Pinyin with tokens in a different manner--- we utilize attention instead of n FC layer.
This choice is primarily influenced by the tokenization strategy of mT5, which uses SentencePiece~\cite{kudo-richardson-2018-sentencepiece}. mT5 tokenizer can group multiple Chinese characters into one single token. For instance, the word ``豆腐'' (meaning ``tofu'') in~\autoref{fig:model} is tokenized as one unit, despite comprising two Chinese characters. Consequently, the corresponding Pinyin representation forms a polysyllabic sequence. We believe that an attention-based approach is more adept at handling such scenarios.

\subsection{Training Strategy}

We design a \emph{three-stage} training recipe for our PmT5 model. 
In the initial stage, we train the model to adapt our newly added structures. This necessitated a continual pre-training procedure on a large-scale text corpus.
In the second stage, our objective is to equip the model with the ability to recognize humor in Chinese sentences effectively. To achieve this, we devise a contrastive learning scheme, which primarily focuses on training the encoder. 
Finally, we fine-tune the model for the generative task using the downstream dataset.

\noindent \textbf{Stage I: Continue pre-training for Pinyin proficiency.}

We collect our pre-training text corpora from a variety of open-source Chinese text datasets, specifically selecting \texttt{wiki2019zh}, \texttt{news2016zh}, and \texttt{baike2018qa} available in the \texttt{brightmart/nlp\_chinese\_corpus} on Github\footnote{\url{https://github.com/brightmart/nlp_chinese_corpus}}.
To enrich the dataset with Chinese humor content, we also incorporate the \texttt{crosstalk} corpus\footnote{\url{https://github.com/FreedomIntelligence/crosstalk-generation}}, which is replete with humor elements from Chinese crosstalk.
Our collective corpora amass more than 11GB, amounting to over 6 billion Chinese characters. This volume surpasses the dataset size used for pretraining in ChineseBERT~\cite{sun2021chinesebert}.
Training our PmT5 (\textbf{PmT5-small}) model with Pinyin input embeddings as extra inputs, we start from the official Hugging Face \texttt{mT5-small} model checkpoint\footnote{\url{https://huggingface.co/google/mt5-small}} and we follow the efficient training recipe of nanoT5~\cite{nawrot2023nanot5}\footnote{\url{https://github.com/PiotrNawrot/nanoT5}}, which enables the pretraining of a T5-base model on the super-large C4 dataset, which is over 750GB, with a time of less than 24 hours using a single NVIDIA A100 GPU.
We leverage the standard masking strategy in pretraining, opting not to use Whole Word Masking (WWM) mentioned in~\cite{cui2021pre} since the default T5 tokenizer inherently tokenizes whole Chinese words into single tokens, thus rendering the application of WWM unnecessary in our context.

\noindent \textbf{Stage II: Contrastive learning for humorous \emph{vs.} normal text discrimination.}

\begin{figure}
    \centering
    \includegraphics[width=\linewidth]{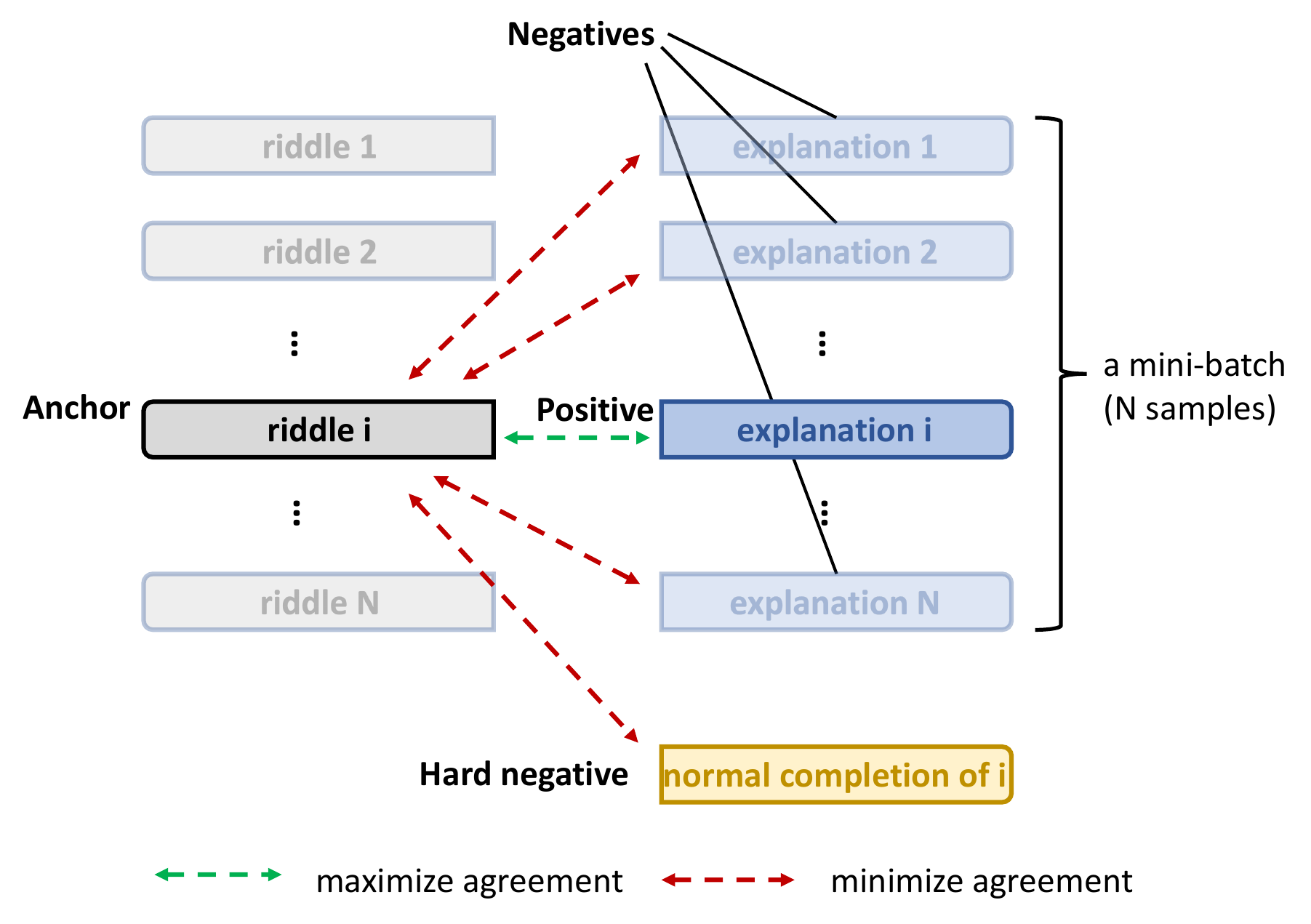}
    \caption{An illustration of the contrastive learning process for allegorical sayings. The semi-circular rectangles represent the output from the encoder. Within each mini-batch, riddle $i$ serves as the \emph{Anchor}, and its correct explanation is identified as the \emph{Positive}, aiming to maximize their alignment. Explanations corresponding to different riddles are considered as \emph{Negatives}, and a normal (not humorous) completion of riddle $i$ which serves as a \emph{Hard Negative}, both intended to minimize agreement with the Anchor.}
    \label{fig:contrastive learning}
\end{figure}

Although the transformer-based mT5 model effectively grasps token-level semantics, it may not adequately capture global semantics at the \emph{sentence level}, which is also crucial for quality sentence generation.
To address this, we employ contrastive learning in our methodology. This technique sharpens the model's ability to differentiate between closely related sentences (positive samples) and those that are less related or unrelated (negative samples), thereby enhancing its grasp of global sentence semantics.
We illustrate our training method in~\autoref{fig:contrastive learning}. Please note that the training happens on the outputs generated by the encoder.
Our contrastive learning strategy is similar to the supervised approach mentioned in SimCSE~\cite{gao2021simcse}. Here, the paired explanation of a given riddle, termed the \emph{anchor}, forms a positive pair with the anchor.
The negative samples for the setup are explanations of all other riddles within the same mini-batch.
Furthermore, to elevate the model's discernment capability on humor versus normal text samples, we introduce an added layer of complexity by incorporating an extra hard negative.
This hard negative is constructed by completing the anchor riddle using an ``\emph{inhumorous}'' seq2seq LM, specifically the default mT5 model in our experiment.
The default mT5 model generates explanations that are technically coherent with the anchor riddle but lack the humorous element typical of allegorical sayings.

Formally, we use $\textbf{r}_i$ and $\textbf{e}_i$ to denote the encoder hidden output of the $i^{\text{th}}$ riddle and explanation in a mini-batch, respectively. We use $\textbf{c}_i$ to denote the normal completion of $\textbf{r}_i$ using the default mT5. The training objective $l_i$ is defined by ($N$ is the size of the mini-batch):

\begin{equation}
    l_i = -\text{log}\frac{\exp(\text{sim}(\textbf{r}_i, \textbf{e}_i)/\tau)}{\exp(\text{sim}(\textbf{r}_i, \textbf{c}_i)/\tau)+\sum_{j=1}^{N}\exp(\text{sim}(\textbf{r}_i, \textbf{e}_j)/\tau)}, 
\label{eq: contra}
\end{equation}
where
\begin{equation}
    \text{sim}(\textbf{r}_i, \textbf{e}_i) = \frac{\textbf{r}_i^\top \textbf{e}_i}{\Vert\textbf{r}_i\Vert \cdot \Vert \textbf{e}_i \Vert}. 
\end{equation}

In~\autoref{eq: contra}, $\tau$ is a temperature hyperparameter.
The difference between~\autoref{eq: contra} and Equation 5 in SimCSE~\cite{gao2021simcse} is that the anchor $\textbf{r}_i$ is only contrasted with the \emph{paired} hard negative $\textbf{c}_i$ instead of all $\textbf{c}_j$ in the mini-batch.

Our contrastive learning and subsequent fine-tuning processes utilize data sourced from the \texttt{chinese-xinhua} corpus\footnote{\url{https://github.com/pwxcoo/chinese-xinhua}}. This corpus comprises clean and authoritative samples extracted from the Chinese Xinhua dictionary (中华新华字典) database, where Xinhua dictionary is a renowned and respected dictionary in China.
Given the critical role of batch size in contrastive learning, we configure it to be 64.

\noindent \textbf{Stage III: Fine-tuning for the downstream task.}

The \texttt{chinese-xinhua} corpus contains $14,032$ instances of allegorical sayings. We randomly shuffle the dataset and partition it into training, validation, and test sets with an 80-10-10 split.
As depicted in~\autoref{fig:demo} (Bottom), we introduce two different generation tasks, with separate model fine-tuning for each task:
\begin{enumerate}
    \item \emph{Completion}: The input of the LM is the riddle, and the target is the explanation.
    \item \emph{Generation from scratch}: The LM's input is the subject of the riddle, while the target is the entire saying. We utilize the HanLP package\footnote{\url{https://pypi.org/project/hanlp/}} to extract the subject.
    To handle instances where multiple riddles map to the same subject, we format the input as ``\{subject\} id'', with ``id'' being a unique numerical identifier for each riddle.
\end{enumerate}

In the fine-tuning process, we explicitly notify the LM about the presence of homophones in the samples. For instance, in the saying ``咸菜烧豆腐——有言（盐）在先'' (Salted vegetables cooked with tofu---have given prior notice (salt)), the homophone ``言（盐）'' (word (salt)), both pronounced ``yán'', is explicitly indicated within \emph{parentheses} to aid in understanding and generation.

\section{A Prompting Approach}
\label{sec:prompting}

Large language models (LLMs)~\cite{openai2023gpt4,touvron2023llama,brown2020language} are known to parameterize vast amounts of world knowledge~\cite{petroni2019language,roberts2020much}. They also exhibit superb in-context learning (ICL) ability~\cite{dong2022survey}, which enables them to effectively learn from provided examples (\aka, demonstrations) and follow natural language instructions (\aka, prompts).
The standout feature of LLMs offers two advantages: (i) the elimination of the need for specific task-oriented learning, and (ii) the ability to ``fine-tune'' the model using interpretable human-written prompts, which makes them \emph{au naturel} LMs for our study.

In this paper, our focus is on examining LLMs' effectiveness considering the two aforementioned generation tasks using \textbf{zero-shot} and \textbf{few-shot} prompting techniques. In the zero-shot scenario, the LLM is provided only a brief description of what a Chinese allegorical saying is, without prior training specifically on these sayings. It's then tasked with completing or generating the saying based on this description alone. In the few-shot scenario, the LLM is additionally presented with a few samples from the \texttt{chinese-xinhua} corpus. The provided samples are randomly sampled from the \emph{training} set.
It is important to note that, although we exclusively use training set samples for LLM to learn the patterns, the potential for data contamination exists. This is due to the availability of our corpora on the web, which might lead to LLMs ``memorizing'' data points from their pre-training dataset~\cite{magar2022data}.

\noindent\textbf{Prompt templates.}

We utilize the following prompt templates outlined in~\autoref{fig:prompt_templates}, originally crafted in \emph{Chinese} and translated here into English to enhance readability for a broader audience.

\begin{figure}[h] %
\centering %

\begin{tcolorbox}[width=\linewidth,colframe=deepblue, sharp corners, colback=white,title={Prompt template for the completion task},coltitle=white,left=1pt,right=1pt,top=1pt,bottom=1pt] 
{
(\textbf{Description}) Chinese allegorical sayings often contain elements of humor or satire, and the beginning and ending should be meaningfully connected.  \\
(\textbf{Demonstrations}) \{examples in the format: riddle---explanation\} \\
Please complete the following Chinese allegorical saying: \\
\{riddle\}
}
\end{tcolorbox}

\begin{tcolorbox}[width=\linewidth,colframe=deepblue, sharp corners, colback=white,title={Prompt template for the generation from scratch task},coltitle=white,left=1pt,right=1pt,top=1pt,bottom=1pt] 
{
(\textbf{Description}) same as above, omitted.\\
(\textbf{Demonstrations})\{examples in the format: Subj: riddle---explanation\} \\
Generate a Chinese allegorical saying based on the following subject: \\
\{subject\}
}
\end{tcolorbox}

\caption{Prompt templates for generating Chinese allegorical sayings.} %
\label{fig:prompt_templates}
\end{figure}

For the above templates, ``\{\}'' are placeholders, and demonstrations are only included in the prompt when using few-shot prompting.
In our experiments, we use ChatGPT (the \texttt{gpt-3.5-turbo-1106} version) as the LLM. The decoding temperature is set to be 0 for consistency.

\section{Experiments and Evaluation}
\label{sec:experiments}

This section evaluates our allegorical saying generation experiment, focusing on its effectiveness using both fine-tuning and prompting techniques. We employ established metrics such as BLEU, ROUGE, and BERTScore for initial assessment. Human evaluators then rate the sayings for coherence and humor, and their feedback is used to train a scoring model. This model effectively measures the quality of the generated content. Key findings are summarized at the end.

\subsection{Assessment Using Established Metrics}

BLEU (Bilingual Evaluation Understudy)~\cite{papineni2002bleu}, ROUGE (Recall-Oriented Understudy for Gisting Evaluation)~\cite{lin2004rouge}, and BERTScore~\cite{zhang2019bertscore} are three commonly used metrics for evaluating the quality of text generated by machine learning models, such as those used in NLP tasks like translation or summarization.

\noindent \textbf{BLEU} is originally used to evaluate the closeness of the machine translation to the human translation (reference).
The core idea is to count the matching n-grams (sequences of words) in the machine translation that appear in the reference translation. The formula for BLEU is as follows:

\begin{equation}
    \text{BLEU} = \text{BP} \cdot \exp\left(\sum_{n=1}^{N} w_n \log p_n\right),
\end{equation}
where $\text{BP}$ is the brevity penalty to prevent overly short translations, $w_n$ are the weights for each n-gram, and $p_n$ are the n-gram precision scores.

\noindent \textbf{ROUGE} is used primarily for evaluating summarization that requires a reference summary. It focuses on the recall.
ROUGE-N calculates the overlap of n-grams between the generated text and the reference text. ROUGE-N is computed as follows:

\begin{equation}
    \text{ROUGE-N} = \frac{\sum_{\text{s} \in \{\text{References}\}} \sum_{\text{gram}_n \in \text{s}} \text{Count}_{\text{match}}(\text{gram}_n)}{\sum_{\text{s} \in \{\text{References}\}} \sum_{\text{gram}_n \in \text{s}} \text{Count}(\text{gram}_n)}.
\end{equation}

ROUGE-L measures the longest common subsequence (LCS) between the generation and references. The formula for ROUGE-L is:

\begin{equation}
    \text{ROUGE-L} = \frac{(1 + \beta^2) \cdot \text{R}_{\text{lcs}} \cdot \text{P}_{\text{lcs}}}{\text{R}_{\text{lcs}} + \beta^2 \cdot \text{P}_{\text{lcs}}},
\end{equation}
where $\text{R}_{\text{lcs}} = \frac{\text{LCS}(X,Y)}{\text{len}(Y)}$ is the recall, with $X$ the generation and $Y$ the reference and $\text{LCS}(X,Y)$ the length of the LCS between $X$ and $Y$.  $\text{P}_{\text{lcs}} = \frac{\text{LCS}(X,Y)}{\text{len}(X)}$ is the precision, and $\beta$ is a hyperparameter to balance between recall and precision.

\noindent \textbf{BERTScore} leverages the pre-trained embeddings from BERT and computes the similarity of the contextual embeddings of words in the candidate and the reference. The BERTScore is computed as follows:

\begin{equation}
    \text{BERTScore} = \frac{1}{|C|} \sum_{c \in C} \max_{r \in R} \text{cosine}(c, r),
\end{equation}
where $C$, $R$ are the sets of token embeddings in the candidate and reference sentence, respectively. And $\text{cosine}(c,r)$ denotes the cosine similarity between two embeddings.

Given the unique characteristics of generative tasks, these metrics are primarily suitable for evaluating completion tasks. This is because, in tasks like generating sayings from scratch based on a subject (as a ``seed''), a reference sequence loses its relevance, rendering traditional evaluation metrics less meaningful.
Evaluation results are showcased in~\autoref{tab:evaluation-task1}. The evaluated text is first tokenized with the \texttt{jieba} package\footnote{\url{https://pypi.org/project/jieba/}}. The values are averaged over the test set.

\begin{table*}[ht]
\centering
\caption{Evaluation of model-generated explanations (TASK 1: completion).}
\label{tab:evaluation-task1}
\begin{threeparttable}
\begin{tabular}{lccccccc}
\toprule
\diagbox{\textbf{Metrics}}{\textbf{Model}\tnote{1}} & \multicolumn{3}{c}{Pre-training and Fine-tuning} & \multicolumn{3}{c}{Prompting} & \emph{Gold}\\
\noalign{\smallskip}
\cline{2-4}
\cline{5-7}
\noalign{\smallskip}
                        & PmT5 (3-stage) & PmT5 (w.o. CL) & mT5 & ChatGPT (ZS) & ChatGPT (FS-5) & ChatGPT (FS-10)  &\\
\midrule
BLEU &\underline{$4.1\times10^{-3}$} &$3.9\times10^{-3}$ &$3.3\times10^{-3}$  &$3.9\times10^{-5}$ &$4.6\times10^{-4}$ &$4.3\times10^{-4}$ &-  \\
ROUGE-1  &\underline{$4.7\times10^{-3}$} &$4.1\times10^{-3}$ &$3.2\times10^{-3}$ &$5.6\times10^{-4}$ &$0$ &$0$ &- \\
ROUGE-2  &\underline{$7.1\times10^{-4}$} &$8.7\times10^{-5}$ &$0$ &$0$ &$0$ &$0$ &- \\
ROUGE-L  &\underline{$4.7\times10^{-3}$} &$4.1\times10^{-3}$ &$3.2\times10^{-3}$ & $5.6\times10^{-4}$ &$0$ &$0$ &- \\
BERTScore  &\underline{0.5865} &0.5849 &0.5812 &0.5647 &0.5769 &0.5751 &- \\
\hline
{\em \textbf{human metrics\tnote{2}:}} \\
coherency &2.13 &2.09 &2.10 &2.38 &\underline{2.45} &2.44 &2.89 \\
humor &\underline{1.59} &1.57 &1.49 &1.27 &1.32 &1.35 &2.06 \\
\bottomrule
\end{tabular}
\end{threeparttable}
\footnotesize
\begin{tablenotes}
   \item{1.} \textbf{PmT5 (3-stage)} is the PmT5 model pre-trained and fine-tuned using the 3-stage strategy. \textbf{PmT5 (w.o. CL)} is the PmT5 model without applying contrastive learning (stage 2), \textbf{mT5} is a standard mT5 fine-tuned on the \texttt{chinese-xinhua} corpus. Both \textbf{PmT5} and \textbf{mT5} refer to the \texttt{small} variants of their respective models. \textbf{ChatGPT (ZS)}, \textbf{(FS-5)}, and \textbf{(FS-10)} represent the ChatGPT prompted using zero-shot prompting, few-shot learning with 5 samples, and few-shot learning with 10 samples, respectively. \textbf{Gold} denotes the gold explanations from the corpus.
   \item{2.} The scores are assigned by the trained scoring model using human annotations. 
\end{tablenotes}
\end{table*}

\begin{table*}[ht]
\centering
\caption{Evaluation of model-generated allegorical sayings (TASK 2: generation from scratch).}
\label{tab:evaluation-task2}
\begin{threeparttable}
\begin{tabular}{lccccccc}
\toprule
\diagbox{\textbf{Metrics}}{\textbf{Model}} & \multicolumn{3}{c}{Pre-training and Fine-tuning} & \multicolumn{3}{c}{Prompting} & \emph{Gold}\\
\noalign{\smallskip}
\cline{2-4}
\cline{5-7}
\noalign{\smallskip}
                        & PmT5 (3-stage) & PmT5 (w.o. CL) & mT5 & ChatGPT (ZS) & ChatGPT (FS-5) & ChatGPT (FS-10)  &\\
\midrule
{\em \textbf{human metrics:}} \\
coherency &2.07 &1.95 &1.89 &2.26 &2.30 &\underline{2.32} &2.89 \\
humor &\underline{1.45} &1.43 &1.42 &0.97 &1.23 &1.22 &2.06 \\
\bottomrule
\end{tabular}
\end{threeparttable}
\end{table*}

\subsection{Human Annotation and Automatic Evaluation}

While metrics like BLEU, ROUGE, and BERTScore provide quantitative evaluations, they often fail to fully grasp the subtleties and cultural nuances inherent in allegorical sayings, especially given their limitations in assessing concise texts. Therefore, it's essential to include human annotators in our evaluation process to address these shortcomings.

We have enlisted two annotators to assess the quality of allegorical sayings, focusing on two crucial dimensions: \textbf{Coherency} and \textbf{Humor}. Each dimension will be evaluated using a scoring scale that ranges from 1 to 3, where 1 indicates low, 2 represents moderate, and 3 signifies high quality.
The two annotators are both proficient at Chinese and are given a detailed scoring guide (human metrics) as follows:

\begin{enumerate}
    \item \textbf{Coherency}:  (1---Low): The saying shows little to no logical connection between its two parts. (2---Moderate): The saying has a somewhat clear connection, but it might require some thought or context to understand fully. (3---High): The saying exhibits a clear and direct logical link between its two parts, making it coherent and easily understandable.
    \item \textbf{Humor}: (1---Low): The saying lacks humor or fails to evoke a sense of amusement. (2---Moderate): The saying is somewhat amusing or witty, but the humor might not be apparent or effective for everyone. (3---High): The saying is distinctly humorous, cleverly employing wit or irony that is likely to amuse a broad audience.
\end{enumerate}

We then randomly sampled 4,000 instances of allegorical sayings for human annotation, comprising 2,000 instances from gold samples, and 1,000 each from PmT5's generations (3-stage) and ChatGPT's generations (few-shot with 5 samples). Each of these instances is scored by \emph{both} annotators. The final human annotation result for each instance will be determined by calculating the mean of the two scores given by the annotators.

\noindent \textbf{Training a scoring model.}

We further fine-tune a \texttt{RoBERTa-wwm-ext-large}\footnote{\url{https://github.com/ymcui/Chinese-BERT-wwm}}~\cite{cui2021pre} model on the supervision of the human-annotated scores. This model is specifically pre-trained for Chinese language understanding. The fine-tuned model then serves as our scoring model for evaluating other generated content. 
Scoring results generated by the fine-tuned model are showcased in the \textbf{\em human metrics} sessions of~\autoref{tab:evaluation-task1} and~\autoref{tab:evaluation-task2}.

\subsection{Findings and Remarks}

Our key findings are summarized as follows:
(i) LMs exhibit the ability to generate allegorical sayings infused with Chinese humor, with our PmT5 model showing superior generation quality due to additional Pinyin input and continued pre-training, as compared to mere fine-tuning a vanilla mT5 on the allegorical saying dataset.
(ii) Leveraging an LLM through few-shot prompting is both user-friendly and effective, yielding results comparable to fine-tuned models, albeit slightly lacking in humor.
(iii) Despite these developments, a marked difference persists in the diversity and depth of sayings generated computationally versus those created through human ingenuity.
(iv) While established NLG metrics like BLEU and ROUGE have limited utility in evaluation, neural-based BERTScore is more effective, yet our scoring model, trained on human annotations, proves to be the most suitable for assessment.
Some illustrative examples are provided in~\autoref{tab:samples}.

\begin{table*}[ht]
\centering
\caption{Some generated examples (explanations).}
\label{tab:samples}
\begin{tabular}{|m{3.7cm}|m{3.7cm}|m{3.7cm}|m{3.7cm}|}
\hline
\diagbox[width=4.1cm]{\textbf{Riddle}}{\textbf{Model Generation}} & \centering PmT5 (3-stage) & \centering ChatGPT (FS-5) & \centering\arraybackslash \emph{Gold} \\
\hline
阎王的蒲扇 (Yama's bushel) & 鬼扯 (Nonsense) & 鬼门关前风和日丽 (A gentle breeze at the gates of hell) & 扇阴风 (Stirring up trouble) \\
\hline
造屋请箍桶匠 (Asking a barrel-maker to build a house) & 找错了人 (Found the wrong person) & 别找错人 (Don't find the wrong person) & 找错人啦 (Got the wrong person) \\
\hline
苍蝇采花 (Flies gathering nectar) & 无孔不入 (Will enter any hole) & 不懂爱情 (Doesn't understand love) & 装疯 (Pretending to be crazy) \\
\hline
王府的管家 (The steward of the royal house) & 管得宽 (Manages broadly) & 门可罗雀 (Sparse as sparrows at the gate) & 欺上瞒下 (Deceiving superiors and subordinates) \\
\hline
稻草人救人 (Scarecrow saving people) & 没道(稻)理 (Makes no sense (rice)) & 也是假的 (Also fake) & 自身难保 (Can't even save itself) \\
\hline
左手喇叭右手鼓 (Trumpet in the left hand, drum in the right) & 想(响)到一块了 (Trying to think (make a sound) together) & 左右开弓 (Shooting arrows with both hands) & 自吹自擂 (Blowing one's own trumpet and beating one's own drum) \\
\hline
\end{tabular}
\end{table*}

\section{Related Work}
\label{sec:related}

\subsection{Computational Humor}

The first line of computational humor research is humor detection, focusing on the automatic decision of a given sentence's humor.
Early efforts centered around statistical machine learning methods such as Naive Bayes and Support Vector Machine (SVM)~\cite{mihalcea2005computational}, as well as the integration of linguistic features~\cite{barbieri2014automatic}.
More recently, this field has evolved to apply transformer-based PLMs for humor detection~\cite{weller2019humor,mao2019bert,annamoradnejad2020colbert}, thanks to their superior NLU capabilities.
Another closely related branch of work is humor generation systems.
In the early years, these systems were predominantly template-based, with lexicons that require human ingenuity~\cite{amin2020survey}, but recent trends include neural generation approaches~\cite{ren2017neural,winters2018automatic,kazi975transformer}. 
The research community has also focused on the generation of question-answer jokes~\cite{binsted1994implemented,ritchie2007practical}, which shares some similarities with our allegorical saying generation.
In Chinese humor research, Wang \etal{}~\cite{wang2021automatic} have explored generating Memes, \ie, short humorous texts that match image content, using models like GPT-2 for captioning.
Moreover, frameworks for Chinese humor evaluation have been proposed~\cite{chen2023can}.

\subsection{Contrastive Learning in NLP}
Contrastive learning is a widely used learning method in the field of unsupervised learning, where the goal is to learn representations by contrasting positive pairs against negative pairs. 
As first introduced by Chopra \etal{} in~\cite{chopra2005learning}, contrastive learning has gained prominence, especially in computer vision~\cite{schroff2015facenet, oord2018representation,he2020momentum,chen2020simple}.
A crucial element of contrastive learning involves generating positive pairs, akin to data augmentation. This process is more challenging in NLP due to the discrete nature of textual data. In NLP, creating semantically similar variations from a given text (the anchor sample) is difficult because of the complexity and variability inherent in language.
A widely used approach to create semantically similar positive pairs is back-translation~\cite{edunov2018understanding}, which involves translating a sentence into another language and then translating it back to the original one, resulting in a semantically similar but syntactically different sentence. 
Techniques from Momentum Contrast (MoCo)~\cite{he2020momentum}, are integrated with back-translation to empower contrastive learning, as demonstrated in~\cite{fang2020cert}.
A notable work of contrastive learning in NLP is SimCSE~\cite{gao2021simcse}, which generates positive pairs through random dropout mask sampling at the latent level, achieving optimal alignment and uniformity in the representation space~\cite{wang2020understanding}.

\section{Conclusion and Future Work}
\label{sec:conclusion}

This paper explores language models' (LMs) proficiency in generating Chinese two-part allegorical sayings, a unique linguistic art form rich in Chinese humor. We employ two methods: pre-training and fine-tuning a smaller LM, and prompting a large language model (LLM). Our modified PmT5 model, based on the mT5, undergoes contrastive learning to distinguish humorous from normal text. Results show both fine-tuned LMs and LLMs can generate quality sayings with Chinese humor, yet they don't quite match human creativity. 

\noindent\textbf{Future work} may involve fine-tuning LLMs or embedding specific cultural-rich Chinese knowledge into smaller LMs for similar tasks. Additionally, exploring the generation of Chinese humor in various modalities could be an interesting avenue.

\bibliographystyle{IEEEtran}
\bibliography{IEEEabrv,ref}

\end{CJK*}
\end{document}